\documentclass[10pt,twocolumn,letterpaper]{article}

\usepackage{cvpr}
\usepackage{times}
\usepackage{epsfig}
\usepackage{graphicx}
\usepackage{amsmath}
\usepackage{amssymb}
\usepackage{microtype}
\usepackage{caption}
\usepackage{subcaption}

\usepackage[pagebackref=true,breaklinks=true,letterpaper=true,colorlinks,bookmarks=false]{hyperref}

\cvprfinalcopy % *** Uncomment this line for the final submission

\def\cvprPaperID{16} % *** Enter the CVPR Paper ID here

\ifcvprfinal\fi
\begin{document}

%%%%%%%%% TITLE
\title{The Northumberland Dolphin Dataset: A Multimedia Individual Cetacean Dataset for Fine-Grained Categorisation \vspace{-6mm}}

\author{Cameron Trotter\and Georgia Atkinson\and Matthew Sharpe\and A. Stephen McGough\and Nick Wright\and Per Berggren\and
Newcastle University\\
Newcastle upon Tyne, UK\\
{\tt\small \{c.trotter2, g.atkinson, m.j.sharpe, stephen.mcgough, nick.wright, per.berggren\}@ncl.ac.uk \vspace{-4mm}}
}

\maketitle

%\thispagestyle{empty}

%%%%%%%%% ABSTRACT
\begin{abstract}\vspace{-2mm}\label{sec:abstract}
Methods for cetacean research include photo-identification (photo-id) and passive acoustic monitoring (PAM) which generate thousands of images per expedition that are currently hand categorised by researchers into the individual dolphins sighted. With the vast amount of data obtained it is crucially important to develop a system that is able to categorise this quickly. The Northumberland Dolphin Dataset (NDD) is an on-going novel dataset project made up of above and below water images of, and spectrograms of whistles from, white-beaked dolphins. These are produced by photo-id and PAM data collection methods applied off the coast of Northumberland, UK. This dataset will aid in building cetacean identification models, reducing the number of human-hours required to categorise images. Example use cases and areas identified for speed up are examined.
\end{abstract}

\vspace{-4mm}
%%%%%%%%% BODY TEXT
\section{Introduction}\vspace{-1.5mm}\label{sec:intro}
As global sea temperatures continue to rise, water pollutants such as plastics increase in number, and greater swathes of coastal areas become urbanised, the monitoring of marine ecosystems is becoming ever more important. Cetaceans (dolphins, whales, and porpoises) make prime candidates for monitoring ecosystem change as they are top predators. They reflect the current state of the ecosystem and respond to change across different spatial and temporal scales \cite{sm08}. Thus there is a significant requirement to monitor individuals within cetacean species to assess population, behaviour and health.

\begin{figure}
\begin{center}
   \includegraphics[width=0.7\linewidth]{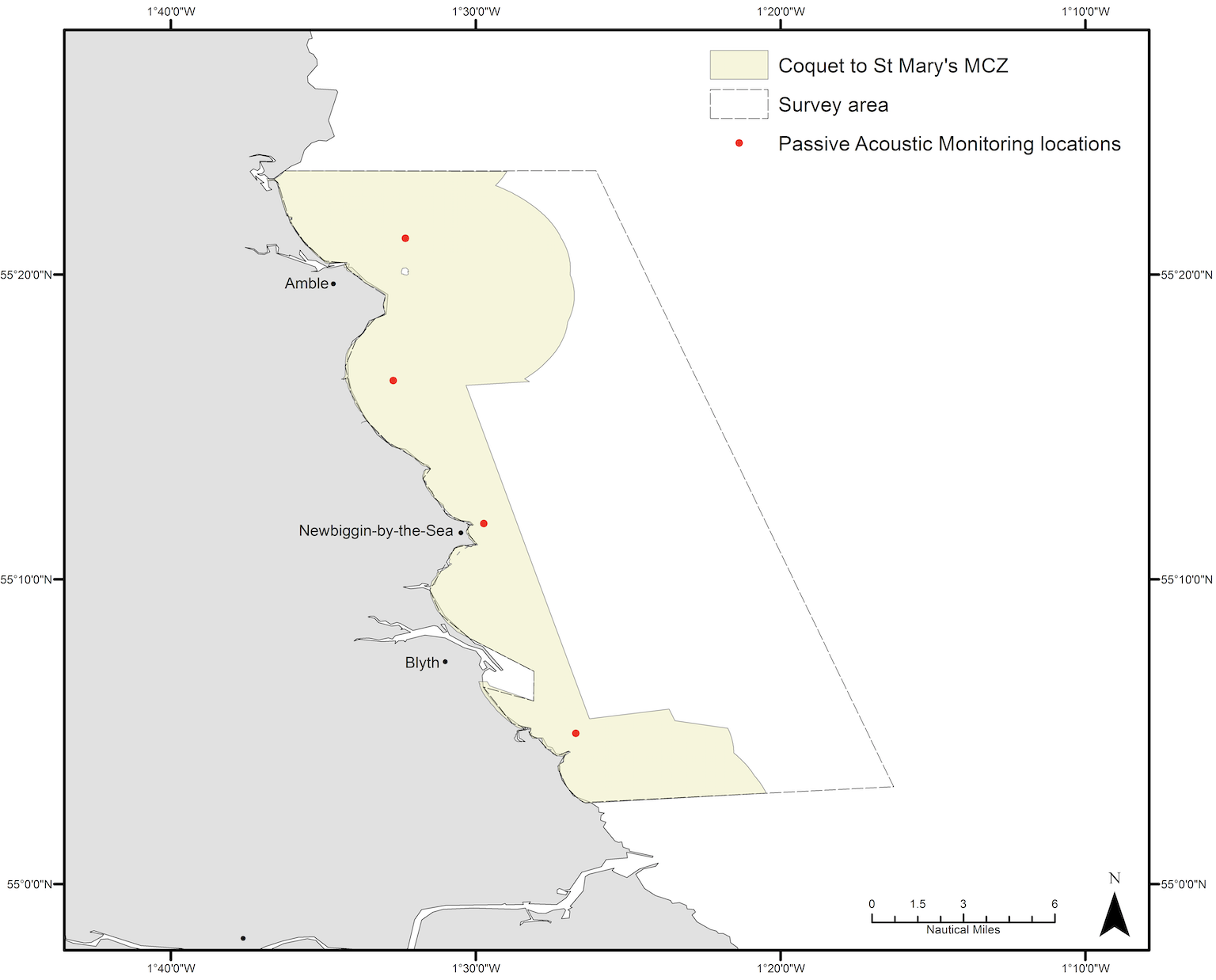}
\end{center}
   \vspace*{-6mm}\caption{Map of the survey area, Northumberland, UK, from St. Mary's Lighthouse in the south to 25nm above Coquet Island in the north.}
\label{fig:map}
\vspace*{-5mm}
\end{figure}

Current methodologies for cetacean research include photo-identification (photo-id) and passive acoustic monitoring (PAM). These methods aggregate large volumes of images and audio which are currently manually investigated. The Northumberland Dolphin Dataset (NDD) is made up of both above and below water images and spectrograms of signature whistles produced from PAM recordings of white-beaked dolphins collected from field work over a number of years around the Northumberland coastline of the UK, see Figure \ref{fig:map}. Both the photo-id and spectrogram images are generated from the same individuals over multiple expeditions. Some individuals appear more than others however, as there is no guarantee that they will be encountered every expedition. The dataset will be discussed in more detail in Section \ref{sec:NDdataset}, along with how it is collected in Section \ref{sec:datacollection}.

The NDD can be utilised for fine-grained categorisation through the use of deep learning techniques, such as Convolutional Neural Networks, which can then be used to aid cetacean researchers. Example use cases for this dataset, that are currently in development, can be seen in Section \ref{sec:examples}.

\subsection{Motivation}\vspace{-1.5mm}

Current identification techniques for cetaceans rely on experts manually identifying individuals. This can often be costly and time consuming due to the number of person-hours required for identification, as well as the large potential for error due to issues such as observer fatigue. Individual identification of dolphins within a species is time consuming due to the nature of the task. Intra-species dolphins have very similar markings and body types making identifying an individual within a pod very difficult. Prominent features must be identified, such as small nicks to the fins, or scars left from injuries to identify an individual. If these features are only prominent on one side of the individual, the task of identification becomes even more difficult. 

With progressively more data being collected during fieldwork through increased use of technology, there is an urgent need for an automatic system for quick identification with reduced error rates. Previous efforts to photo-id individuals from underwater video stills took around three months from raw video file to completely catalogued. It is hoped the NDD will provide a springboard for computer vision researchers to develop fine-grained systems to aid in the speed up and reduction of errors.

%-------------------------------------------------------------------------
%-------------------------------------------------------------------------
\section{The Northumberland Dolphin Dataset}\vspace{-1.5mm} \label{sec:NDdataset}

\begin{figure}
\begin{center}
   \includegraphics[width=\linewidth]{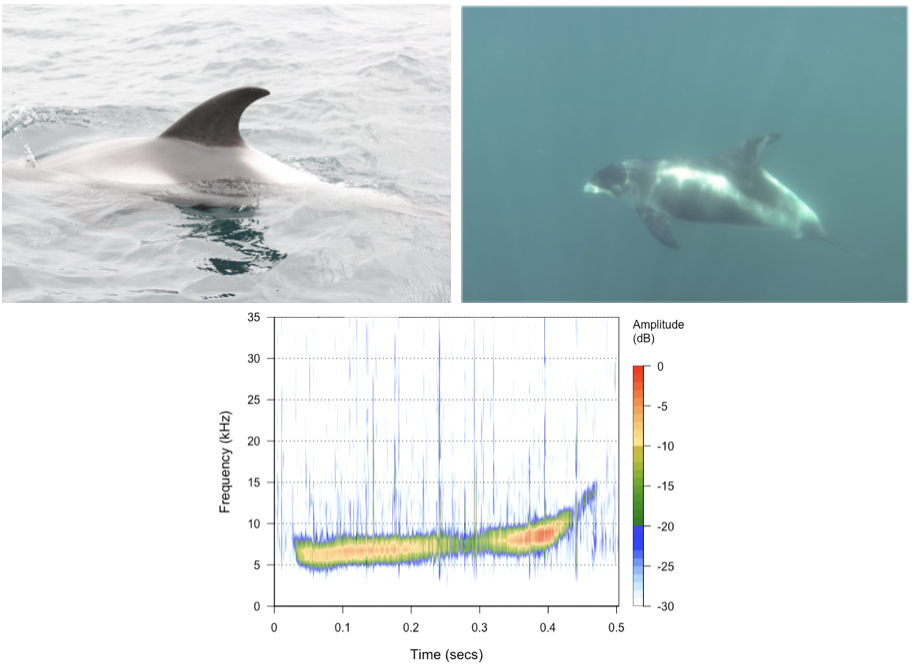}
\end{center}
   \vspace*{-5mm}\caption{Examples of the images found in the NDD. \textit{Top left}: photo-id above-water, \textit{top right}: photo-id below-water, \textit{bottom}: signature whistle spectrogram.}
   \vspace*{-6mm}
\label{fig:examples}
\end{figure}

The NDD consists of two subsets, each representing two methodologies used for cetacean identification, photo-id and PAM, called the photo-id dataset and the signature whistles dataset respectively. Each of these contain differing images which can be used by computer vision researchers to develop methodologies of automatic identification, utilising the same data as what would be available to marine biologists performing identification manually. Examples of the images can be seen in Figure \ref{fig:examples}. Currently, the NDD contains 6649 images gathered from previous small scale data collection efforts and 2 hours 40 minutes of audio recordings from which spectrograms have been created. It is expected that this collection will grow to around 26000 images and 24 hours of audio after the 2019 fieldwork season, as a concerted data collection effort is being made to improve and expand the dataset. This estimate is based off similar fieldwork conducted in Zanzibar, Tanzania in 2015 over a 3 month time frame for a separate project. It is expected that around 30-40 expeditions will be undertaken in the 2019 fieldwork season. 

%-------------------------------------------------------------------------
\subsection{Photo-id Images}\vspace{-1.5mm}\label{sec:photo-id}

The photo-id dataset consists of both above and below water images of white-beaked dolphins found in the survey area -- Figure \ref{fig:map}. Above-water images are large panoramic photographs taken from a vessel during an encounter. These images contain at least one fin, potentially multiple if a pod is present. Marine biologists use prominent markings on the fin, such as scars and lesions, as well as fin shape, for identification. For below-water images, there can be one or multiple individuals contained in the image. Identification is achieved using prominent markings on any visible body surface. 

%-------------------------------------------------------------------------
\subsection{Signature Whistle Spectrograms}\vspace{-1.5mm}\label{sec:whistles}

The signature whistle dataset consists of spectrogram images of whistles produced by audio recordings of white-beaked dolphin vocalisations. A hydrophone is placed into the water attached to buoy at the beginning of encounters and moved if the dolphins drift too far from the hydrophone. The whistles -- often referred to as signature whistles -- are then extracted from the audio recordings and saved individually. Signature whistles are defined as whistles with unique frequency curves, produced by individual dolphins within a pod. As with the photo-id dataset, multiple whistles produced from different individuals can be present in the audio recording at any one time, leading to possibility of whistles being seen as overlapping within a spectrogram. Identification is achieved using the whistle shapes produced by the spectrograms as there is evidence to support white-beaked dolphins produce signature whistles.  

%-------------------------------------------------------------------------
\subsection{Data Challenges}\vspace{-1.5mm}\label{sec:challenges}

\begin{figure}
\begin{center}
   \includegraphics[width=0.8\linewidth]{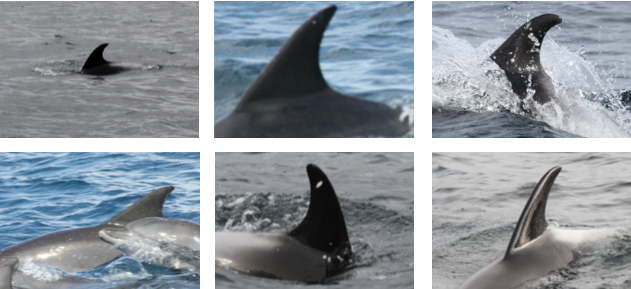}
\end{center}
   \vspace*{-5mm}\caption{Examples images of common challenges from top left to bottom right. (1) \textbf{Region of Interest too small}. (2) \textbf{Blur}. (3) \textbf{Splash}. (4) \textbf{Partial fin}: The cetacean fin may be partially obstructed. (5) \textbf{One sided markings}: Useful identifiers may only be present on one side. (6) \textbf{Angle of approach}.}
\label{fig:photo-id-challenges}
\vspace*{-4mm}
\end{figure}

Both datasets provide their own unique challenges. The photo-id dataset, images have been included which are difficult even for experienced marine biologists to identify. Six common issues with above-water cetacean photo-id, which are present in the dataset, have been identified and can be seen in Figure \ref{fig:photo-id-challenges}. Complexities for below-water dataset images are similar, with the added issue of water conditions affecting visibility of markings. Visibility conditions are not an issue with the above-water images as surveys are not conducted during adverse weather conditions. However, underwater images allow for identification based on more parts of the body, such as the tail stock or ventral, increasing the chances of identification.

\begin{figure}
    \begin{center}
    \begin{subfigure}[b]{0.49\linewidth}
        \centering
        \includegraphics[width=0.9\linewidth]{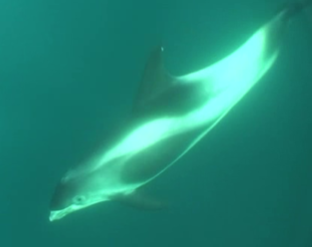}
    \end{subfigure}%
    \begin{subfigure}[b]{0.49\linewidth}
         \centering
        \includegraphics[width=0.9\linewidth]{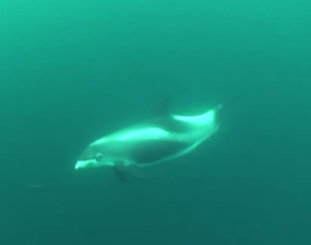}
    \end{subfigure}
    \end{center}
    \vspace*{-5mm}\caption{Two individuals in the dataset may be very similar looking to each other. \textit{Left}: individual F11001, \textit{right}: individual F11002.}
    \label{fig:notTwins}
    \vspace*{-5mm}
\end{figure}

Furthermore, a large number of individuals in the photo-id dataset are visually very similar, see Figure \ref{fig:notTwins}. It is not a simple task to identify individuals; even if at first the identification of dolphin individuals may be thought of as being in some way similar to human facial recognition.

The signature whistles dataset also contains challenging images. Within this data, due to levels of underwater noise, some whistles do not show up clearly when produced into spectrograms, even if they can be heard clearly in the audio recordings. Ongoing field work will enable labelled data to collected by the use of underwater video and audio recording taken in the same place. This will allow for loud whistles produced to be matched with dolphins in video that are close to the camera. 

Both the photo-id and signature whistles data are unbalanced. There are largely varying number of samples per individual. This again adds complexity to the dataset, providing an interesting challenge when identifying lesser-seen individuals.

%-------------------------------------------------------------------------
%-------------------------------------------------------------------------
\section{Data Collection}\vspace{-1.5mm} \label{sec:datacollection}

Data collection surveys are conducted along the Northumberland coast, UK, from St Mary’s Lighthouse to North of Coquet Island (approx. 25 nautical miles (nm)) and cover an area to up to 6nm offshore. This survey area includes the Coquet to St Mary's Marine Conservation Zone (MCZ). A map of the survey area can be seen in Figure \ref{fig:map}. In a typical season, approximately 40 surveys across three months are conducted, restricted to days of fair weather and sea state \textless 3 on the Beaufort scale\cite{world1970beaufort}.

Surveys systematically cover the area following designated line transects using a 5.6m rigid-hulled inflatable boat. Observers scan for groups of dolphins using the continuous scanning method \cite{mann1999behavioral}. Above-water images from the photo-id dataset are collected using a camera from the vessel, whilst below water images are captured using multiple Go-Pro cameras fixed to the vessel's hull below the waterline. The signature whistles dataset is collected from hydrophones in the water around the vessel. The collected audio recordings are processed using Goertzel's algorithm to detect whistles converted into spectrograms after basic band-pass filtering. Images collected from sightings by other recreational water users are also included.

%-------------------------------------------------------------------------
%------------------------------------------------------------------------
\section{Example Experiments}\vspace{-1.5mm} \label{sec:examples}

The NDD will allow marine biologists and computer vision researchers to work collaboratively. Reducing identification time whilst allowing for novel fine grained image categorisation research.

Prominent marker identification is one example use case for NDD. Using a set of images of the same identified individual, prominent features can be recorded and used for identification in the future. These features may be found on only one individual, for example a pattern or set of scars, or may be multiple markers which on their own cannot identify an individual but given their arrangement in a certain orientation can.

Individual identification is another example use case. Using either unseen above or below water images, whistle spectrogram, an individual can be identified. This can be in the form of top-5 suggestions and confidence scores displayed to the biologist to aid their manual process, greatly reducing the set of fins a human identifier needs to observe to find a positive match. If no images of the individual exist in the catalogue, then the system suggests the assignment of a new identifier.

Finally, abundance estimation could also be performed. Using spectrograms produced from an encounter, individuals can be identified and also new unseen whistles can be determined. Using this information, abundance estimates can be made by counting the number of individuals identified, and the number of new distinct whistles. This will allow biologists to catalogue this information for every encounter to aid their research. 
%-------------------------------------------------------------------------
%-------------------------------------------------------------------------
\section{Conclusion}\vspace{-1.5mm}\label{sec:conclusion}

Thanks to advances in computer vision it is now within the realm of possibility not to just perform species identification, but identify on a more fine-grained level down to the individual. Coupled with the increasing use of technology in the marine conservation area, and the now vast amount of data being collected during fieldwork, it is of the utmost importance that methodologies are developed to reduce the number of person-hours spent by biologists performing data wrangling, especially at a time of heightened ecological awareness and desire to undertake conservation. The NDD provides computer vision researchers a fine-grained dataset on which to develop solutions to the problems described, whilst also facilitating interdisciplinary work between the fields of computing and marine biology. It is believed this dataset is the first to provide individual above and below water labelled cetaceans, along with spectrogram imagery of signature whistles produced by them. 

Work is ongoing to improve the dataset, by providing labels to the spectrograms, which will allow for the identification of individuals using these images. Future work on the dataset will involve providing bounding boxes for individuals in the above and below water images. This will be of extreme value in the event of multiple individuals in the same image, providing location to individuals labelled. 

%-------------------------------------------------------------------------
{\small
\bibliographystyle{ieee_fullname}
\bibliography{NON_Anon_Paper}
}

\end{document}